\documentclass{article}

\usepackage[preprint]{spconfa4}
\usepackage{array}
\usepackage{booktabs}
\usepackage{cite}
\usepackage{graphicx}
\usepackage{multirow}
\usepackage{epsfig}
\usepackage{amsmath,amssymb}

\newcommand\etal{\emph{et~al.}}

\begin{document}

\title{Document Layout Analysis via Dynamic Residual Feature Fusion}

\name{Xingjiao Wu$^{1,2}$, Ziling Hu$^{2}$, Xiangcheng Du$^{2,3}$, Jing Yang$^{2*}$\thanks{$^*$Corresponding author.}, Liang He$^{1,2}$}
\address{\small{1 Shanghai Key Laboratory of Multidimensional Information Processing, East China Normal University, Shanghai, China}\\
\small{2 School of Computer Science and Technology, East China Normal University, Shanghai, China}\\
\small{3 Videt Tech Ltd., Shanghai, China}\\
{\small \{52184506007, 51184506059, 51194506062\}@stu.ecnu.edu.cn, \{jyang, lhe\}@cs.ecnu.edu.cn}
\thanks{Xingjiao Wu and Ziling Hu contributed equally to this work.}
}
\maketitle

\begin{abstract}
The document layout analysis (DLA) aims to split the document image into different interest regions and understand the role of each region, which has wide application such as optical character recognition (OCR) systems and document retrieval.
However, it is a challenge to build a DLA system because the training data is very limited and lacks an efficient model.
In this paper, we propose an end-to-end united network named Dynamic Residual Fusion Network (DRFN) for the DLA task.
Specifically, we design a dynamic residual feature fusion module which can fully utilize low-dimensional information and maintain high-dimensional category information.
Besides, to deal with the model overfitting problem that is caused by lacking enough data, we propose the dynamic select mechanism for efficient fine-tuning in limited train data.
We experiment with two challenging datasets and demonstrate the effectiveness of the proposed module.

\end{abstract}


\section{Introduction}
\label{sec:intro}

Document layout analysis (DLA) is an important research area dedicated to extracting semantic information from the document image.
 As a critical preprocessing step of document understanding systems, DLA can provide information for several applications such as document retrieval, content categorization, and text recognition.

Traditional DLA methods are highly dependent on manually designed features and can not deal with complex layouts.
 Recently, with the rapid development of deep learning, significant progress has been made in DLA research. Inspire by Fully Convolutional Network(FCN)~\cite{long2015fully}, many deep learning-based methods are proposed. These methods consider DLA as a special segmentation task~\cite{chen2017convolutional}.
However, it is still a challenge to design an efficient DLA system. Accurately predicting the category of regions in a document is limited by the bottlenecks that lack enough training data and a suitable end-to-end model. This work focuses on implementing a universal and effective DLA framework in which an effective network and an appropriate update learning method are designed.

Because the current dataset size can not satisfy the training of the generalized model, many researchers dedicated to using synthetic data to help the training model\cite{yang2017learning}. However, it is still a difficult problem that there is a distribution gap between synthetic data and real data.
Inspired by the work of~\cite{guo2019spottune,li2019selective}, we propose a dynamic select mechanism, an approach for efficient input-related fine-tuning. The core of the approach is that we split a layer into a pre-train path and a fine-tune path and aggregate information from both paths to generate two channel-wise select weights for aggregating the feature maps. With a slight increase in trainable parameter, the dynamic select mechanism can prevent the model from overfitting.
%Moreover, we adopted an iteratively update strategy with a changeable loss scale factor to quickly adapt model to real data.
\begin{figure}
	\centering
	\includegraphics[width=1\linewidth]{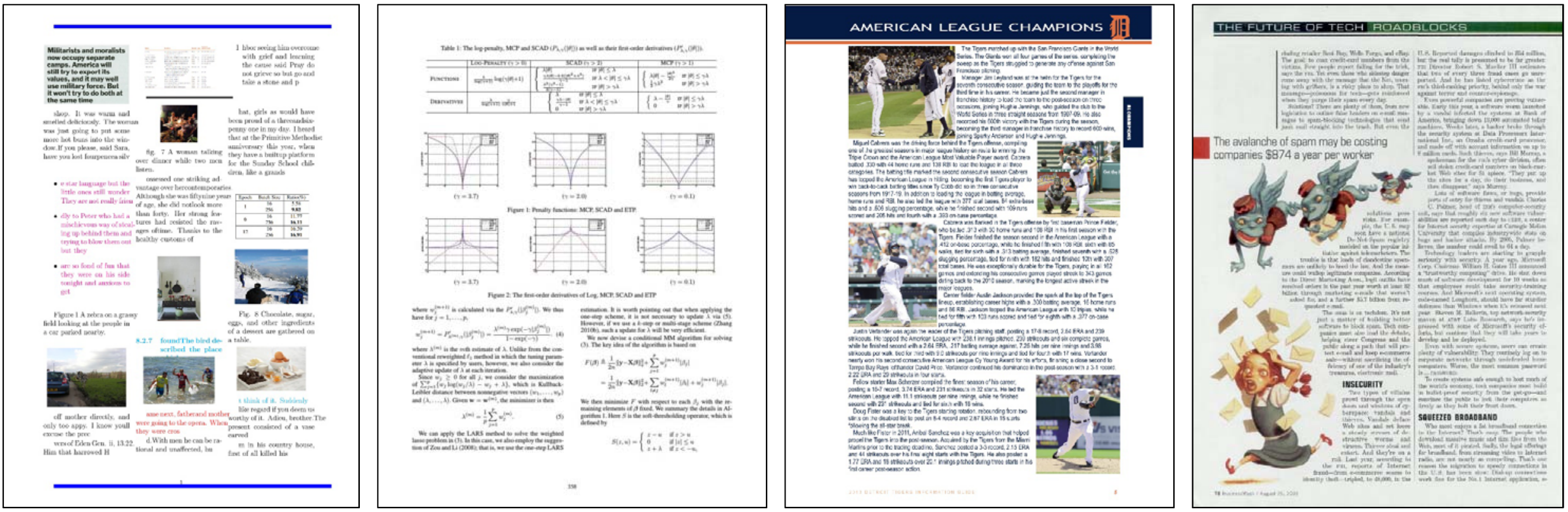}
	\caption{Images from different sources. The image in column 1 is synthetic data and others are from the real dataset.}
	\label{Fig6}
\vspace{-3px}
\end{figure}

The dynamic selection mechanism provides a new model encoding method, but the final model also needs to be decoded.
How to transfer the previous information most effectively is also an important issue.
Many works have discussed high-dimensional semantic information and low-dimensional semantic information. High-dimensional features are abundant in category semantic information and low-dimensional features skilled in reconstructing original resolution detail.
Both features are important to accurate pixel classification.
To utilize all dimensional features, the FCN model directly superimposes features extracted from the encoder layers on the decoder layers. However, this mechanism is harmful to the category of semantic information in high-dimensional features.
In this work, we designed a dynamic residual feature fusion module that effectively fuses the different dimension features and generates robust fusional features. The core idea of the proposed dynamic residual feature fusion module is to extra global semantic information from high-dimensional features and weight fusional features as guidance. Then high-dimensional features are added to fusional features to maintain high-dimensional category semantic information flow.

Our contributions can be summarized as follows:

\begin{itemize}
	\item We propose a dynamic residual feature fusion module which can effectively fuse the high-dimensional category semantic and low-dimensional detail information for document layout analysis.
	\item We propose a dynamic select mechanism that can prevent the model from overfitting when fine-tuning in limited data.
	\item The DRFN achieves comparable results on two benchmark datasets.
\end{itemize}

\section{Related Work}
\label{sec:related}

\begin{figure*}
\centering
	\includegraphics[width=0.92\linewidth]{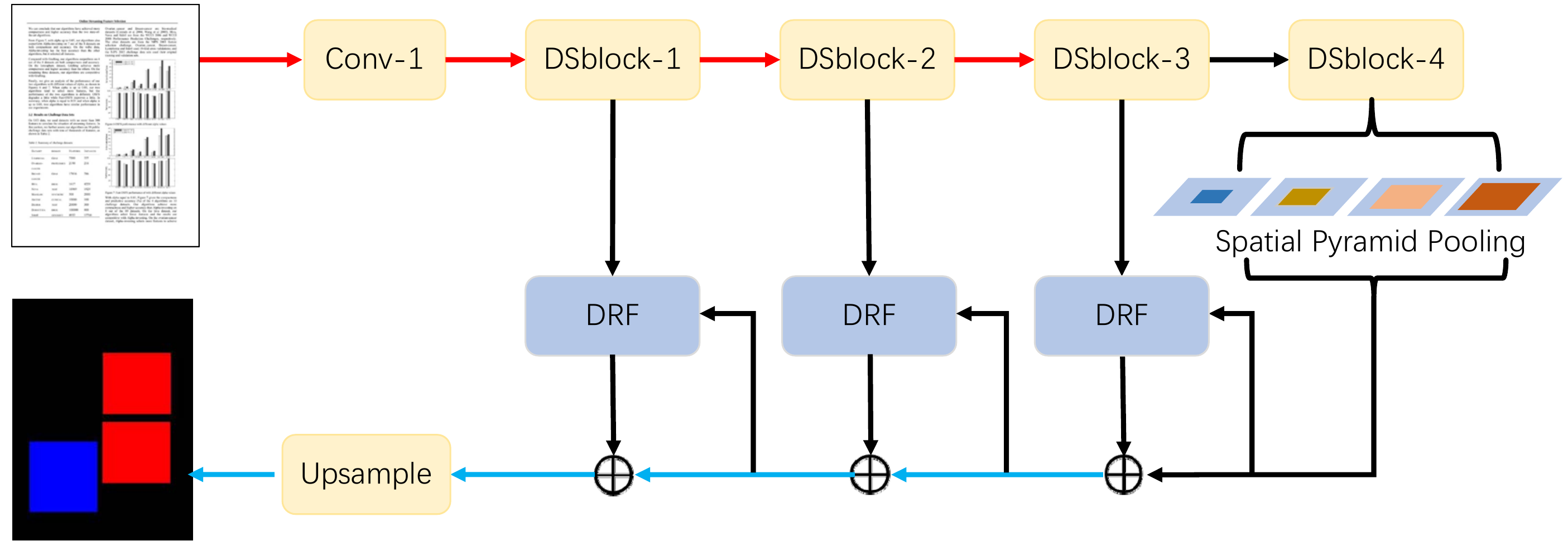}
	\caption{Illustration of DRFN. We adopt the encoder-decoder architecture and use Resnet-18 as the backbone. The top half is the encoder part. The DSblock is a resblock with a dynamic select mechanism. We use a mechanism on each resblock. The lower half is the decoder part. We adopt the dynamic residual feature fusion module as the decoder block. The red and blue lines represent the downsample and upsample operators respectively.}
	\label{Fig1}
\end{figure*}

In the early stage, the classical document layout analysis (DLA) strategies can be roughly divided into two categories: bottom-up and top-down. Usually, the bottom-up method obtains document analysis result dynamically from a smaller granularity data level. In general, bottom-up analysis strategies use local features from documents such as pixel distributions and properties of connected components to detect individual pixels or words. Then, detected elements are grown up to form larger document regions and stop until they reach the presenting analysis level. The bottom-up strategies have attracted researchers because they have high accuracy and can deal with complex layouts. However, most methods are high time complexity and require large space. The classical bottom-up strategies can be subdivided into five core categories: connected component analysis\cite{o1993document}, texture analysis \cite{journet2005text}, learning-based analysis\cite{marinai2005artificial}, Voronoi diagrams\cite{kise1998segmentation}, and Delaunay triangulation\cite{eskenazi2015delaunay}.

The top-down strategies iteratively divide whole pages into small document elements such as columns, blocks, textiles, and words. Similarly, the top-down strategies can be subdivided into four categories: Texture-based Analysis\cite{saabni2014text}, Run Length Smearing Algorithm (RLSA)\cite{alaei2011new}, DLA projection-profile\cite{shafait2010effect}, and With space analysis\cite{shafait2008background}. Usually, the top-down strategies work greatly on regular document layouts. However, it is limited by the requirement for clean and skews corrected document images. Though it can deal with regular common rectangular document layouts, the bottom-up strategies and the top-down strategies are not effective enough to handle complex layouts.

With the rapid development of deep learning, serious methods are proposed and achieve impressive performance in this field. The DLA task is considered as an especially semantic segmentation task\cite{ravi2016semantic}, whose goal is performing a pixel-level understanding of the segmentation objects. Deep learning-based DLA methods are mainly inspired by Fully Convolutional Network (FCN)\cite{long2015fully}. He~\etal~ \cite{he2017multi} uses an FCN with the multi-scale feature for semantic page segmentation with an auxiliary element contour detection task. Wick~\etal~ \cite{wick2018fully} adopt a modified Fcn to deal with historical document segmentation. Li~\etal~ \cite{li2018deeplayout} propose a novel end-to-end deep neural network named DeepLayout for page layout analysis.

The low-dimensional features contain more detailed information and the high-dimensional features contain more semantic information\cite{chen2013blessing}. For generating more robust features, FCN fuse the low-dimensional features and high-dimensional features using a skip connection structure. However, the skip connection structure can not take advantage of the high-dimensional features, because the contained high-dimensional category semantic information is confused by low-dimensional features. Therefore, we propose a dynamic residual feature fusion module. This module achieves recovering images detail and meanwhile maintaining the category semantic information.

\section{Framework}
\label{sec:app}

In this section, we will introduce the details of our methods. Firstly, we elaborate on the overall architecture of the proposed network. Then, the dynamic residual fusion feature module is expounded. Next, we present the details of the dynamic select mechanism.
Finally, we will introduce synthetic document data.

\subsection{Network Architecture}

The overview of the proposed network is shown in Fig~\ref{Fig1}. We adopt the idea that trains an FCN network extracting feature for pixel-wise segmentation with dynamic residual feature fusion module and dynamic select mechanism. For extracting features effectively, we adopt an encoder-decoder structure, which is widely used in general image semantic segmentation task. To build a powerful encoder, we apply ResNet-18 as the backbone and adopt dynamic select mechanism on each restock. Moreover, the dilated convolution with the rate of 2 is applied to DSBlock-4, so the output size of feature maps from the encoder is 1/16 of the input image. Then, we use spatial pyramid pooling proposed in Deeplabv3+~\cite{chen2017deeplab} to enhance the feature \textsl{F} output from the encoder. After that, we upsample \textsl{F} and feed it in the decoder part. We perform the dynamic residual feature fusion module as a fast and effective decoder structure, which can maintain high-dimensional category information while making full use of low-dimensional detail information.

\subsection{Dynamic Residual Feature Fusion Module}

The encoder-decoder networks mainly consider using the encoder to extract category information and recover spatial information by the decoder. PSPNet\cite{zhao2017pyramid} and Deeplab\cite{chen2017deeplab} directly use bilinearly upsample, which are disadvantageous to restoring the spatial position to the origin resolution because of lacking low-dimensional detail information. FCN \cite{long2015fully} and PANet~\cite{li2018pyramid} gradually recover clear object boundary details by using different scale low-dimensional features. However, these methods directly combine different scale features. It may be harmful for category semantic information in high-dimensional features and cause inaccurate pixel classification. Therefore, we propose a dynamic residual feature fusion module which maintains the category semantic information by a residual struct. Besides, global semantics in high-dimensional can be a guidance weight to select precise detailed information.

\begin{figure}
	\centering
	\includegraphics[width=0.9\linewidth]{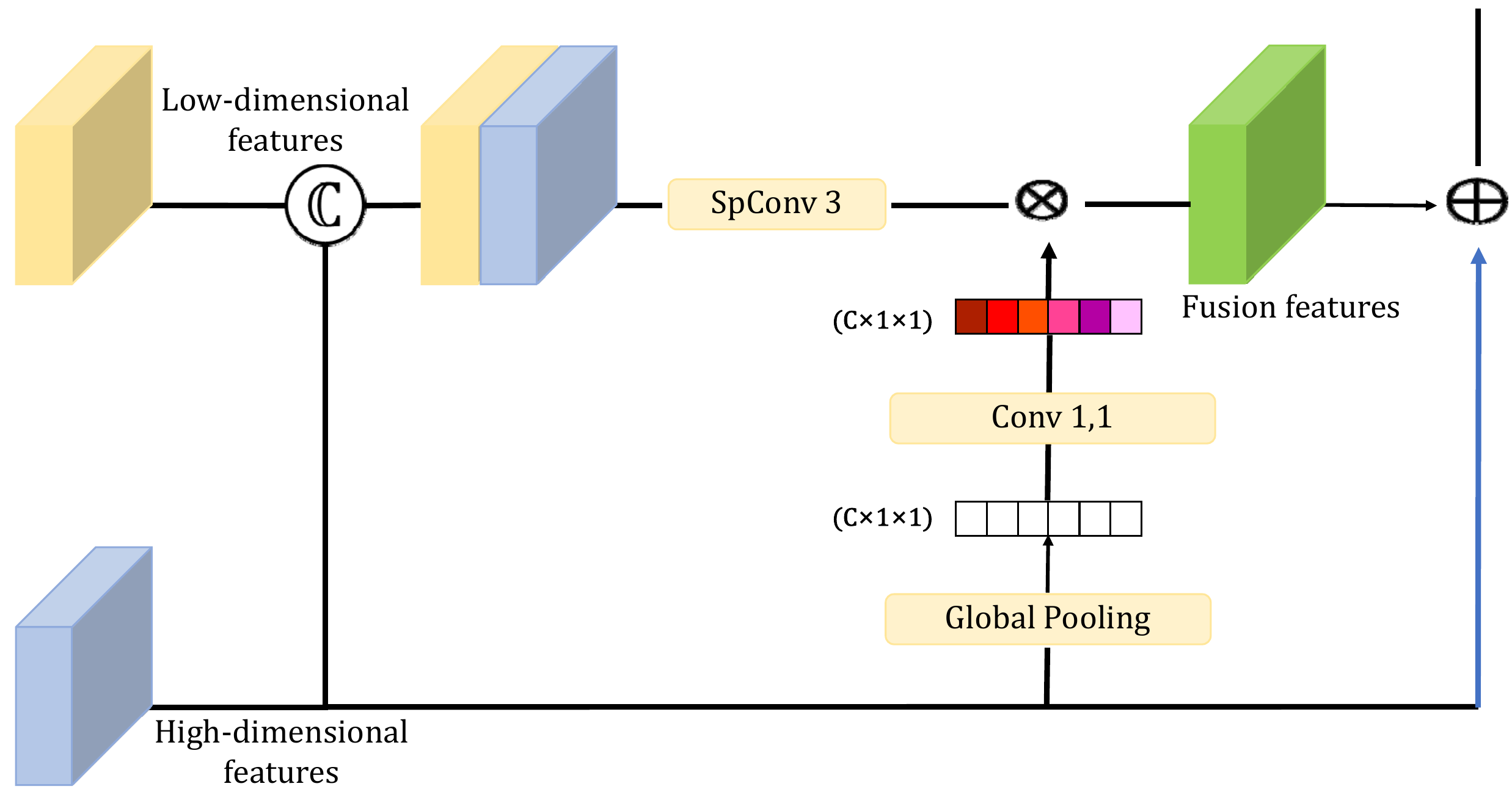}
	\caption{Dynamic Residual Feature Fusion Module Structure. Where 3 represent 3$\times$3 convolutional layer and 1 represents 1$\times$1 Convolutional layer. SpConv represents depthwise separable convolution. The blue line represents upsample operators.}
	\label{Fig2}
\end{figure}

Our dynamic residual feature fusion module is shown in Fig~\ref{Fig2}. In detail, we first concatenate the low-dimensional features and high-dimensional features. Then we deploy a 3$\times$3 convolution layer to reduce the channel size and generate the fusion features. For reducing calculation overhead, we adopt the depthwise separable convolution \cite{chollet2017xception}. Then the guidance weight is extracted from high-dimensional features through a global average pooling and two 1$\times$1 convolutional layers with batch normalization and sigmoid activation. The first convolutional layer compact the size of guidance weight with a reduction ratio r (we set as 16 in our experiments), and the second one restore it to its original size. Then the guidance weight is multiplied by the fusion features. Finally, the high-dimensional features are added with the fusion features. This module maintains the category information flow in the decoder layer and deploys different dimension features.

\subsection{Dynamic Select Mechanism}
\begin{figure}
	\centering
	\includegraphics[width=1\linewidth]{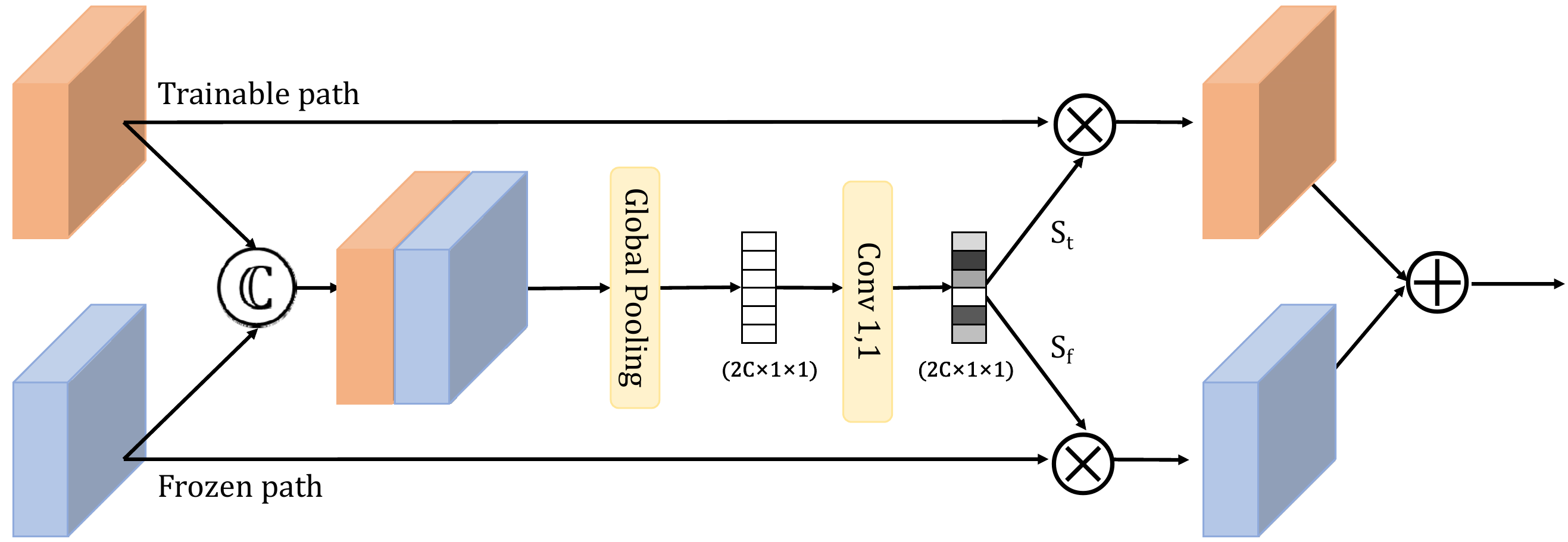}
	\caption{Trainable path and frozen path in dynamic select mechanism structure.}
	\label{Fig3}
\end{figure}
Deep learning-based vision model has shown remarkable success in many computer vision fields but often rely on large amounts of the labelled training examples. It is a challenge that trains a universal document layout analysis model because of lacking enough data. To deal with this problem, much research adopts a pre-trained model trained in the large synthetic dataset and fine-tune the model in the target dataset. However, fine-tuning in a small dataset may result in overfitting. Therefore, we proposed the dynamic select mechanism.

Given a pre-trained network model, our dynamic select mechanism split the encoder block as two parallel trainable fine-tune paths $P_{t}$ and frozen per-trained path $P_{f}$. Then our mechanism dynamically generates two channel-wise select weights \textsl{$S_t$} and \textsl{$S_f$}. We implement our method on Resnet\cite{he2016deep}, the output of the $l-th$ encoder block is computed as:

\begin{equation}\label{eq1}
x_{l} = S_{t}(x_{l-1})P_{t}(x_{l-1}) + S_f(x_{l-1})P_{f}(x_{l-1}) + x_{l-1}
\end{equation}
Where $S_{t}(x_{l-1})$ $\in$ $\mathbb{R}^{C\times
	1\times1}$ and $S_{f}(x_{l-1})$ $\in$ $\mathbb{R}^{C\times
	1\times1}$.

For generating the two select weights, we need to gather information from both parallel paths. We first fuse the results from two paths through a concatenate operation. Then global statistics are extracted via a global average pooling.
Next, similar to the dynamic residual feature fusion module, two 1$\times$1 convolutional layers are adapted to generate the global weight \textsl{S}. Then we reshape \textsl{S} to the size of $\mathbb{R}^{C\times 2\times1}$ and apply a softmax function. Finally, \textsl{$S_t$} and \textsl{$S_f$} are obtained by splitting \textsl{S}. On each channel, the sum of \textsl{$S_t$} and \textsl{$S_f$} is always equal to 1.

Our model dynamically decides a channel-wise feature select strategy for each image to fully utilize both towpaths during the whole training process.

\section{Experiments}
\label{sec:exp}

To evaluate the effectiveness of the DRFN, we conducted several sets of experiments. We use DSSE-200 and CS-150 as benchmark datasets. We train DRFN on 4000 synthetic data for pre-train and fine-tune in two real benchmark datasets. Besides, we use DSSE-200 to verify the effect of the dynamic residual feature fusion module and dynamic select mechanism. To measure the performance on DLA task, we divide the classification criteria into four classification criteria: background, figure, text, table. We use accuracy, precision, recall and F1 metrics to evaluate the DRFN.

\subsection{Datasets}

%\noindent
\textbf{DSSE-200. } The DSSE-200\cite{yang2017learning} is a complex document layout dataset including various dataset styles. The dataset contains 200 images from pictures, PPT, brochure documents, old newspapers and scanned documents.

\noindent
\textbf{CS-150. } The CS-150\cite{clark2015looking} was proposed by Clark. The dataset consists of 150 papers and includes 1175 images. The classification criteria in this dataset are divided into three types: image, table, and others.

\subsection{Implementation Details}
The baseline is the fully convolutional neural network (FCN). We use Resnet-18 pre-trained on Imagenet\cite{deng2009imagenet} as the backbone of the DRFN because it has excellent performance and lightweight parameters. We use Adma as the optimizer, and the batch size is set to 8.
For the data argument, we adopt several strategies. Firstly, the images are horizontally flipped with the probability of 0.5. Then we flip up the images with the probability of 0.5. Next, we randomly crop samples with a size ratio of 0.7 from the transformed images.

For training our model, we generate 4000 synthetic images as the per-train dataset. When fine-tuning, we adopt the dynamic select mechanism. We first pre-train our model on the per-train dataset. Then we use the pre-trained parameter to initialize the model. During the fine-tuning process, we use real data from each dataset.
Because the DSSE-200 and CS-150 do not provide standard training datasets, we randomly select some images as the train data and others as test data. We selected 40 images on DSSE-200 and 940 images on CS-150.

\begin{figure}
	\centering
	\includegraphics[width=0.92\linewidth]{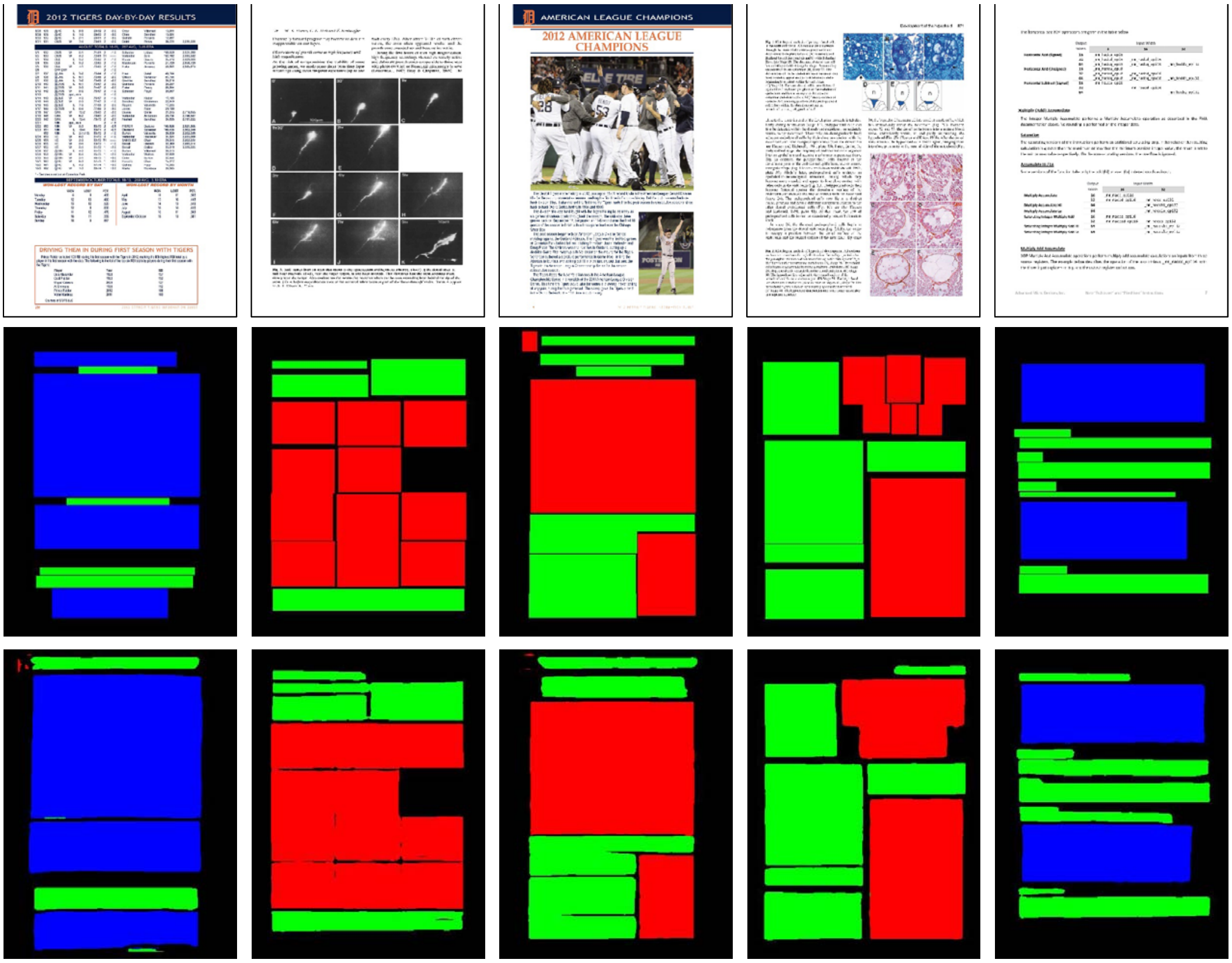}
	\caption{ The DSSE-200 real documents and their corresponding segmentation. Top: original. Middle: ground-truth. Bottom: predictions. Segmentation label colors are: figure (red), table (blue) and background (black).
	}
	\label{Fig5}
\end{figure}

\begin{table}
	\centering
	\caption{The results on the DSSE-200 dataset. The baseline is FCN/Resnet-18.}
	\begin{tabular}{clcccc}
      %\begin{tabular}{p{30px}<{\centering}p{80px}p{20px}<{\centering}p{20px}<{\centering}p{20px}<{\centering}p{20px}<{\centering}}
		\toprule
		Model   &Method                           &A                &P            &R             &F1\\ \midrule
		1    	&FCN/VGG16\cite{long2015fully}    &80.6             &71.0         &77.6          &74.1\\
		2   	&FCN/Resnet-50                    &81.2             &75.7         &75.4          &75.5\\
		3   	&baseline                         &80.7         	&72.2         &81.6          &76.6\\
		4   	&DRFN  		                      &82.8             &77.1         &83.7          &80.3\\  \bottomrule

	\end{tabular}
	\label{tabel1}
\end{table}

\subsection{Ablation Study}
\label{sec:Ablation}

To verify the effectiveness of our approach, we do a series of comparative experiments on the DSSE-200 dataset. These experiments mainly focus on evaluating dynamic residual feature fusion module and dynamic select mechanism.
%\vspace{0.08in}

% \usepackage{booktabs}
\begin{table}[]
\small
\caption{The result on dataset. The DV3+ means the DeeplabV3+ model with Xception backbone. The DV3+R18  means DV3+/Resnet-18.  The Para represents the trainable parameter amount of the model.}
\label{tabel2}
\begin{tabular}{@{}p{35px}p{9px}<{\centering}p{9px}<{\centering}p{9px}<{\centering}p{11px}<{\centering}|p{9px}<{\centering}p{9px}<{\centering}p{9px}<{\centering}p{9px}<{\centering}p{9px}<{\centering}p{9px}<{\centering}@{}}
\toprule
\multicolumn{1}{c}{\multirow{2}{*}{Method}}                  & \multicolumn{4}{c}{DSSE-200} & \multicolumn{4}{|c}{CS-150} & \multicolumn{1}{c}{\multirow{2}{*}{Para}}     \\ \cmidrule(l){2-9}
 \multicolumn{1}{c}{}                                        & A     & P     & R     & F1   & A     & P    & R    & F1   &  \multicolumn{1}{c}{}   \\  \midrule
Segnet~\cite{badrinarayanan2017segnet}   & 87.0  & 82.5  & 86.0  & 84.2 & 99.4  & 93.8 & 94.1 & 94.0 & 29M  \\
PSPnet~\cite{zhao2017pyramid}            & 88.4  & 87.7  & 83.5  & 85.5 & 99.3  & 94.0 & 93.5 & 93.8 & 46M  \\
PANet~\cite{li2018pyramid}               & 87.9  & 86.0  & 82.4  & 84.2 & 99.3  & 94.2 & 94.0 & 94.1 & 168M \\
DV3+~\cite{chen2018encoder}              & 87.0  & 83.3  & 81.7  & 82.4 & -     & -    & -    & -    & 53M  \\
DV3+R18                                  & 90.2  & 87.6  & 87.1  & 87.3 & 99.4  & 94.6 & 94.5 & 94.6 & 15M  \\
baseline                                 & 87.1  & 83.3  & 83.8  & 83.5 & 99.4  & 94.0 & 94.1 & 94.1 & 12M  \\
DRFN                                     & \textbf{90.5}  & \textbf{88.8}  & \textbf{90.1}  & \textbf{89.5} & \textbf{99.5}  & \textbf{95.3} & \textbf{94.8} & \textbf{95.1} & 17M  \\ \bottomrule
\end{tabular}
\end{table}

\noindent
\textbf{Dynamic Residual Feature Fusion}
For evaluating the effectiveness of dynamic residual feature fusion module, we compared our model with a baseline on DSSE-200 dataset. Here we directly deploy Resnet-18 on as the encoder, without dynamic select mechanism. And we only use synthetic data to train all models. We also evaluate the performance with a different backbone. The results are shown in Table\ref{tabel1}. Compared with the baseline, FCN\cite{long2015fully} with Resnet-18, our model achieved a better performance in all metrics. Besides, an interesting first phenomenon can be seen by comparing model 2 and model 3. Though with a deeper and more parameters backbone, the model does not achieve better performance and even make a degradation. This is the reason why we chose Resnet-18 as the backbone.

\begin{figure}
	\centering
	\includegraphics[width=.92\linewidth]{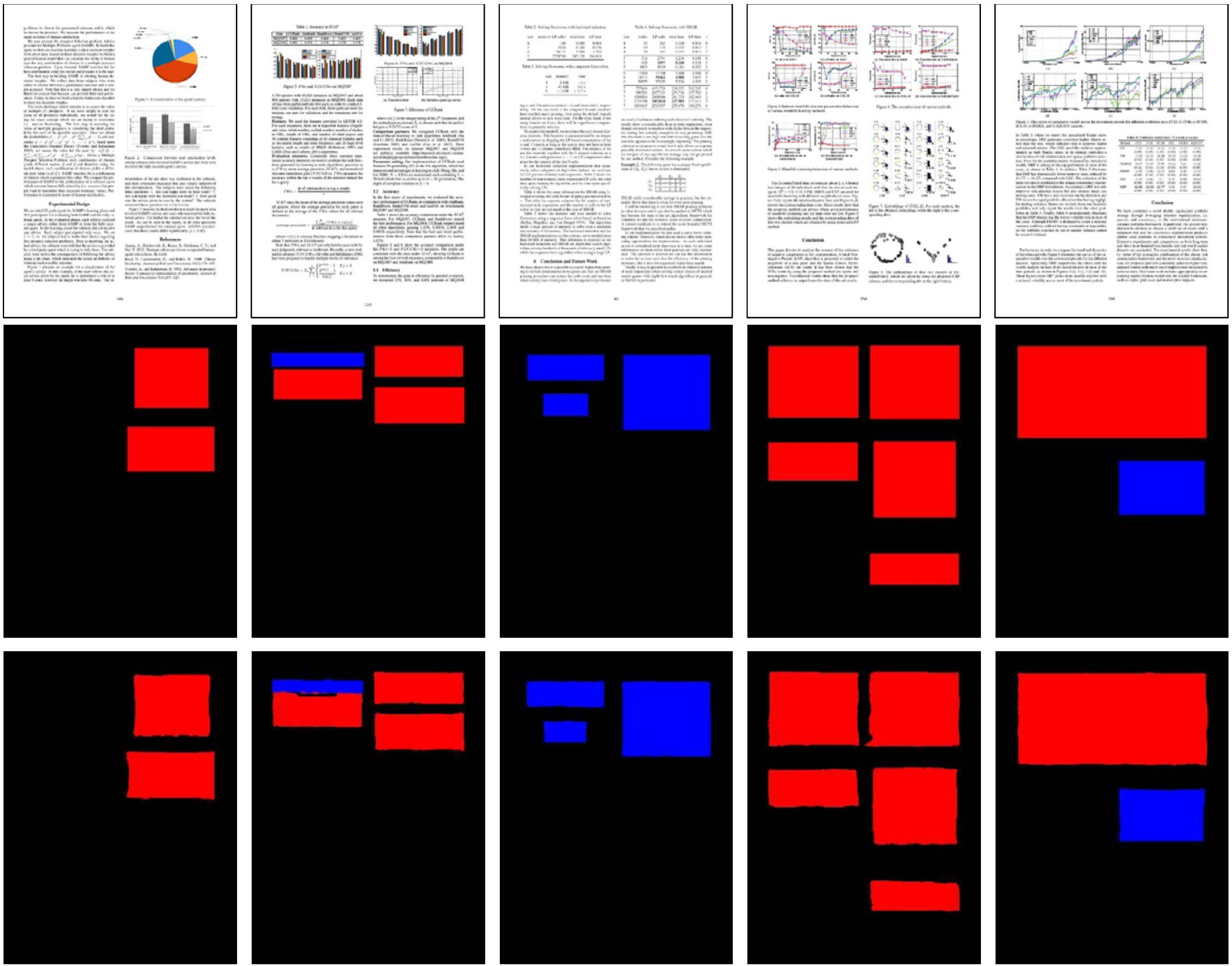}
	\caption{ The CS-150 real documents and their corresponding segmentation. Top: original. Middle: ground-truth. Bottom: predictions. Segmentation label colors are: figure (red), table (blue) and background (black).
	}
	\label{Fig4}
\end{figure}

\vspace{0.06in}
\noindent
\textbf{Dynamic Select Mechanism}
To evaluate the effectiveness of the dynamic select mechanism, we fine-tune the model on DSSE-200 dataset and evaluate the performance. The results are shown in Table\ref{tabel4}. In our implementation, we initialize the model with the parameter pre-trained on synthetic data. Compared with the model directly fine-tune with Resnet-18, dynamic select mechanism improved model performance on all metrics.

\vspace{0.06in}
\subsection{Comparisons with Prior Arts}
\textbf{Comparisions on DSSE-200 dataset} (Table\ref{tabel2}). Following the setting mentioned above, we experiment on the DSSE-200. Compared with other classical semantic segmentation methods, the experiment shows that our models are more powerful with much less parameters.  Moreover, as shown in Fig~\ref{Fig5}, our model can correctly identify text pixels which are unlabeled in the groud-truth. This result proves our model adapts the target data while retaining the knowledge learned in the source synthetic data.

\vspace{0.06in}
\noindent
\textbf{Comparisions on CS-150 dataset} (Table\ref{tabel2}). We also evaluate our model on the CS-150 dataset. Because only the image area and the table area are marked in CS-150 dataset, we process the label to classify the image as background, figure, and table. The results are shown in Fig~\ref{Fig4}. Observing the experimental results, our results reach the sate-of-the-art.

\begin{table}
	\centering
	\caption{The fine-tune results on the DSSE-200 dataset. The DSM represents the dynamic select mechanism.}
	\begin{tabular}{lcccc}
		\toprule
		Method                          &A                &P            &R             &F1\\ \midrule	
		fine-tune with Resnet-18        &90.3             &88.3         &88.5          &88.4\\		
		fine-tune with DSM  		    &90.7             &88.8         &90.1          &89.5\\ \bottomrule
		
	\end{tabular}
	\label{tabel4}
\vspace{-8px}
\end{table}

\section{Conclusions}
\label{sec:conclusion}

In this paper, we propose a novel solution for constructing a model of universal document layout analysis.
To make full use of low-dimensional information and maintain high-dimensional semantic information, we propose the dynamic residual feature fusion module.
Furthermore, we propose the dynamic select mechanism for efficient input-related fine-tuning.
Extensive experiments on document layout analysis benchmarks show the superior performance of the proposed method.

\bibliographystyle{IEEEtran}
\bibliography{total}

\end{document}